\documentclass[letterpaper]{article} 
\usepackage[draft]{aaai2026}  
\usepackage{times}  
\usepackage{helvet}  
\usepackage{courier}  
\usepackage[hyphens]{url}  
\usepackage{graphicx} 
\usepackage{natbib}  
\usepackage{caption} 
\usepackage{algorithm}
\usepackage{algorithmic}

\usepackage{newfloat}
\usepackage{listings}
\DeclareCaptionStyle{ruled}{labelfont=normalfont,labelsep=colon,strut=off} 
\floatstyle{ruled}
\newfloat{listing}{tb}{lst}{}
\floatname{listing}{Listing}
\usepackage{pifont}
\usepackage{multirow}
\usepackage{amsmath}
\usepackage{subcaption}

\usepackage{adjustbox}
\usepackage{tikz, pgf}
\usetikzlibrary{arrows}
\usetikzlibrary{positioning}

\newcommand\smalldash{\raisebox{0.5ex}
                       {\begin{tikzpicture}
                        \protect\draw[->, dashed] (0,0) -- (0.5,0);
                        \end{tikzpicture}}}

\newcommand\dotted{\raisebox{0.5ex}
                    {\begin{tikzpicture}
                     \protect\draw[->, dotted] (0,0) -- (0.5,0);
                    \end{tikzpicture}}}

\title{Taxonomy-Agnostic Neurosymbolic Grounding of Robot Kinematics}

\author{
    Bastien Dussard,
    Aurélie Clodic,
    Guillaume Sarthou
}
\affiliations{
    LAAS-CNRS, Université de Toulouse, Toulouse, France\\
    \{bastien.dussard, aurelie.clodic, guillaume.sarthou\}@laas.fr
}

\begin{document}

\maketitle

\begin{abstract}
While symbolic reasoning for task allocation requires grounding a robot's physical capabilities, standard Unified Robot Description Format (URDF) files provide only low-level kinematics without the formal semantic context needed for reasoning. We present \texttt{URDF2Ontology}, an automated, taxonomy-agnostic neurosymbolic framework that transforms raw URDF files into populated, grounded OWL ontologies.
It leverages a Large Language Model (LLM) for targeted semantic reasoning, coupled with an iterative \textit{sequential graph-purging} mechanism to prevent context window saturation. To process complex kinematic structures, it employs a morphology-aware dual-track architecture featuring a bottom-up \textit{Tree Track} for branching assemblies and a top-down \textit{Chain Track} for serial limbs. Guided by neurosymbolic guardrails, the framework extracts functional sub-graphs and asserts populated ontologies aligned with arbitrary user-provided taxonomies. Empirical evaluation across a diverse fleet of heterogeneous robots demonstrates strong schema alignment, valid structural extraction, and a significant reduction in manual domain authoring overhead.
\end{abstract}

\begin{links}
   \link{Code}{https://github.com/RIS-WITH/urdf2ontology}
\end{links}

\section{Introduction}

As autonomous robot fleets scale across heterogeneous applications, semantic interoperability is critical for high-level task allocation, capability matching, and automated planning. Fleet management systems rely on symbolic representations of platform capabilities (e.g., mobility, end-effectors, sensors) to select suitable hardware for specific tasks. Establishing a flexible semantic layer eliminates reliance on hard-coded planning domains, allowing fleet architectures to scale without exhaustive manual knowledge engineering.

Recent robotic architectures leverage Large Language Models (LLMs)~\cite{brohan2023can}, Vision-Language Models (VLMs)~\cite{driess2023palm}, and Vision-Language-Action (VLA) models~\cite{zitkovich2023rt2} to map instructions into plans or actions. However, critical bottlenecks persist: they struggle to generalize across heterogeneous physical embodiments and incur high computational overhead in multi-robot fleets.
Nevertheless, while fleets cannot rely solely on foundation models for real-time control, LLMs excel at targeted semantic reasoning. Recent frameworks exploit LLMs' implicit commonsense to extract structured properties~\cite{adamik2025extracting} and populate 3D scene representations~\cite{zemskova20253dgraphllm}, bridging continuous perception and abstract symbolic knowledge.

However, while environment grounding has been extensively studied, embodiment grounding remains neglected. In knowledge representation, foundational frameworks like KnowRob~\cite{beetz2018know} and SOMA~\cite{bessler2021foundations} capture formal mappings between hardware and functional capabilities. Yet, populating these ontologies across diverse platforms remains a manual, labor-intensive bottleneck. Crucially, existing pipelines are rigidly bound to fixed, proprietary schemas, lacking the flexibility to map hardware onto arbitrary user taxonomies.

This bottleneck stems from standard description formats like the Unified Robot Description Format (URDF). Designed for kinematic modeling, URDF files represent hardware as raw trees of links and joints without semantic annotations. While formats like Semantic Robot Description Format (SRDF) offer manual semantic overlays, component identifiers in raw URDFs are often non-descriptive or arbitrarily nested, preventing rule-based parsers from mapping hardware onto formal taxonomies.

To address this, we present \texttt{URDF2Ontology}, a neurosymbolic framework that automatically translates URDF models into formal, ontology-aligned semantic representations. Given a raw URDF and a target taxonomy, our system generates a populated knowledge base mapping low-level morphology to high-level concepts. By coupling a multi-pass abstraction workflow with constrained local LLM queries, the framework transforms non-semantic descriptions into verified knowledge structures without manual authoring, fine-tuning, or cloud inference.

The primary contributions of this paper are as follows: 1) The \texttt{URDF2Ontology} framework: a multi-pass neurosymbolic pipeline utilizing sequential graph purging to abstract \textbf{raw URDF files into populated Description Logic ontologies}.
2) A novel structural alignment strategy combining bottom-up assembly for branching topologies (sensors, end-effectors, locomotion) with top-down decomposition for continuous chains (body parts), \textbf{adaptable to arbitrary user-supplied schemas}.
3) Integration of local LLM constraints for more robust and context-efficient reasoning, while \textbf{ensuring alignment with the target ontology}.

\section{Related Work}

\subsection{Robot Embodiment in Knowledge Representation}

Foundational efforts have been made in the knowledge representation field to develop precise and re-usable abstractions of robot's symbolic needs for decision-making. 

\paragraph{Ontological Modeling:} The IEEE 1872 standards establish a comprehensive knowledge-representation framework where CORA~\cite{prestes2013towards} defines upper robotics domain, RTR~\cite{balakirsky2017towards} represents task execution workflows, and AuR~\cite{gonccalves2021ieee} models autonomous behavior. While those ontologies tackle critical aspects of automated reasoning for robot systems, they primarily target high-level concepts and standardized task semantics rather than physical hardware configurations. These frameworks specify what a robot is conceptually, where it operates, and what goals it can execute, yet they lack explicit concepts for a robot's actual hardware embodiment. Other ontologies were later proposed to bridge this gap, such as KnowRob~\cite{tenorth2017representations} or SOMA~\cite{bessler2021foundations}. While allowing for more granular embodiment representation and their link with functional capabilities, they still require manual knowledge engineering work to create the mapping between low-level robot description (e.g., URDF) and high-level standard ontologies. 

\paragraph{Automated Population:} 

Automating robot morphology population has been previously explored through conceptual modeling of the URDF taxonomy~\cite{kunze2011towards, borcard2025towards}. 
For instance, the Semantic Robot Description Language (SRDL)~\cite{kunze2011towards} automatically imports kinematic models into an ontology aligned with URDF schema primitives (e.g., Link, Joint). While this approach establishes a 1:1 mapping between raw URDF elements and ontology instances, it fails to ground elements into high-level component taxonomies. For example, it translates \textit{gripper\_joint} into a generic prismatic joint rather than linking it to a functional gripper concept. Consequently, establishing semantic hardware classifications still requires manual assertion.

\subsection{Automated LLM-Grounding in Robotics}

To bypass fragile rule-based parsers and the manual overhead of knowledge engineering, recent work leverages LLMs for neurosymbolic grounding, translating perception and hardware artifacts into formal symbolic representations.

\paragraph{Semantic Environment Grounding:}

Most robotic semantic grounding focuses on the operational environment. Previous work demonstrated that LLMs can extract commonsense object and environmental properties from parametric weights to populate ontologies~\cite{adamik2025extracting} or translate Universal Scene Descriptions (USD) into grounded scene graphs via zero-shot prompting by exploiting structural and lexical cues~\cite{shuai2026usd}. However directly applying zero-shot prompting to raw URDF remains problematic. Unlike static scene graphs, URDFs encode interdependent kinematic topologies with unstandardized frame nomenclature and non-physical dummy frames. Providing an undecomposed tree in a single prompt frequently leads to topological hallucinations and reasoning failures across multi-hop joint-link dependencies.

\paragraph{Semantic Embodiment Grounding:}

Embodiment grounding enables task planners to adapt execution strategies across diverse hardware. For instance, systems like EMOS~\cite{chen2025emos} leverage LLMs to synthesize high-level "embodiment resumes" from URDF, allowing an LLM-based task planner to match platform capabilities (e.g., mobility, perception, manipulation) to specific mission constraints. 
While this approach extracts a functional abstraction, it bypasses formal modeling in favor of unstructured natural language summaries. It thus obfuscates the robot's compositional architecture, severing the explicit links between low-level hardware components and high-level capability semantics~\cite{dussard2023ontological}. Without component-level symbolic abstraction, systems cannot verify physical sub-component constraints, track structural modifications, or interface cleanly with formal Description Logic reasoners.
\\ 

In summary, current approaches force a trade-off between manually populated ontologies that support rigorous compositional reasoning and automated LLM extractions that lack formal symbolic guarantees. \texttt{URDF2Ontology} bridges this gap by combining topology-aware extraction with taxonomy-guided LLM queries, enabling automated, verified semantic grounding from raw URDFs into formal ontologies.

\section{Framework Overview}

In this section, we define the mapping problem between raw URDF models and target semantic taxonomies. We then outline the macro-level multi-pass workflow before introducing our running example. Finally, we describe the cross-cutting neurosymbolic guardrails that underpin the pipeline.

\subsection{Problem Formulation}

Standard URDFs bundle high-level kinematic topologies with low-level physics descriptors (e.g., visual/collision mesh files, inertia matrices, etc). In practice, these numerical attributes contribute high token entropy without offering meaningful semantic for classification, frequently leading to LLM context saturation and hallucinations. To overcome this, we operate on a purely structural kinematic graph, defined as a stripped directed acyclic graph $\mathcal{G}_{\text{URDF}} = (\mathcal{V}_L, \mathcal{E}_J)$:

\begin{itemize}
  \item $\mathcal{V}_L = \{l_1, l_2, \dots, l_n\}$ is the set of link nodes representing rigid body identifiers.
  \item $\mathcal{E}_J = \{j_1, j_2, \dots, j_m\}$ is the set of joint edges defining parent-child relative kinematic linkages.
\end{itemize}

We formulate our semantic target in Description Logic (DL) using the standard TBox/ABox formalism $\mathcal{O} = \langle \mathcal{T}, \mathcal{A} \rangle$. The Terminological Box $\mathcal{T} = (\mathcal{C}, \mathcal{R})$, provided to the framework by the domain expert, defines:

\begin{itemize}
  \item $\mathcal{C}$: The set of domain concepts and taxonomic hierarchies (e.g., $\text{Camera} \sqsubseteq \text{VisionSensor} \sqsubseteq \text{Sensor}$).
  \item $\mathcal{R}$: The set of allowed object properties (e.g., $\text{partOf}$, $\text{isAttachedTo}$, $\text{hasCapability}$).
\end{itemize}

The Assertional Box $\mathcal{A}$ is the populated output synthesized by the framework. It consists of:

\begin{itemize}
  \item $\mathcal{I}$: The set of named individuals extracted from link nodes ($l_i \in \mathcal{V}_L$) and aggregated subgraphs.
  \item Concept assertions $C(i)$, where $i \in \mathcal{I}$ and $C \in \mathcal{C}$.
  \item Role assertions $R(i, j)$, where $i, j \in \mathcal{I}$ and $R \in \mathcal{R}$.
\end{itemize}

Rather than hardcoding a rigid, proprietary taxonomy into the pipeline, \texttt{URDF2Ontology} is designed to be taxonomy-agnostic. The core objective is to dynamically align the structural entities of an arbitrary URDF with the conceptual space of any valid input TBox $\mathcal{T}$ (e.g., KnowRob, CORA, SOMA, or custom user ontologies). This means that given a stripped kinematic graph $\mathcal{G}_{\text{URDF}}$ and an user-supplied schema $\mathcal{T}$, the framework has to populate an ABox $\mathcal{A}$ such that every generated individual $i \in \mathcal{I}$ and assertion in $\mathcal{A}$ is strictly aligned with the axioms of $\mathcal{T}$.

\subsection{The Multi-Pass Workflow}

Attempting to translate an entire URDF graph into a semantic ontology in a single monolithic pass leads to performance degradation due to token clutter, deep topological nesting, and semantic interference between distinct functional modules. 
To overcome this, \texttt{URDF2Ontology} executes an iterative, concept-ordered Outer Loop that relies on domain-specific \textit{a priori} knowledge of robot morphology: terminal leaf nodes predominantly correspond to sensors; end-effectors and locomotion systems may host sensors (but not vice versa); and all functional components connect back to the root link through structural body parts.
The Outer Loop thus consists of four sequential passes:

\begin{itemize}
  \item \textbf{Pass 1 (Sensors):} Sensors typically reside at terminal branches or outer attachments. Isolating and classifying them first eliminates fine-grained frame noise.
  \item \textbf{Pass 2 (End-Effectors):} Terminal manipulators (e.g., grippers, tools) are identified at the extremities of remaining kinematic sub-trees.
  \item \textbf{Pass 3 (Locomotion):} Mobile bases, steerable wheel assemblies, or leg attachment points are extracted to ground the robot's mobility modality.
  \item \textbf{Pass 4+ (Body Parts):} With peripheral attachments removed, the remaining structure is processed to extract primary structural kinematic chains (e.g., arms, head).
\end{itemize}

During pass $k$, after the pipeline grounds a target subgraph $\mathcal{G}' \subset \mathcal{G}_{\text{URDF}}^{(k-1)}$ and populates the ABox, $\mathcal{G}'$ is systematically \textbf{purged}. 
Purging removes the links and joints encompassed in the abstracted sub-graphs from the active URDF graph, thus updating it to show new terminal leaf links.

\begin{equation*}
\mathcal{G}_{\text{URDF}}^{(k)} = \text{Purge}\left(\mathcal{G}_{\text{URDF}}^{(k-1)}, \mathcal{G}'\right)
\end{equation*}

In this manner, following Pass 1, all sensor-related links are abstracted and removed, producing a simplified graph $\mathcal{G}_{\text{URDF}}^{(1)}$ where $|\mathcal{V}_L^{(1)}| \!<\! |\mathcal{V}_L^{(0)}|$. Subsequent passes (e.g., end-effector abstraction) operate on a significantly reduced search space, effectively preventing token window saturation and eliminating cross-domain context pollution.

\subsection{The PR2 Running Example}

\begin{figure}[!h]
\centering
\includegraphics[width=0.88\columnwidth]{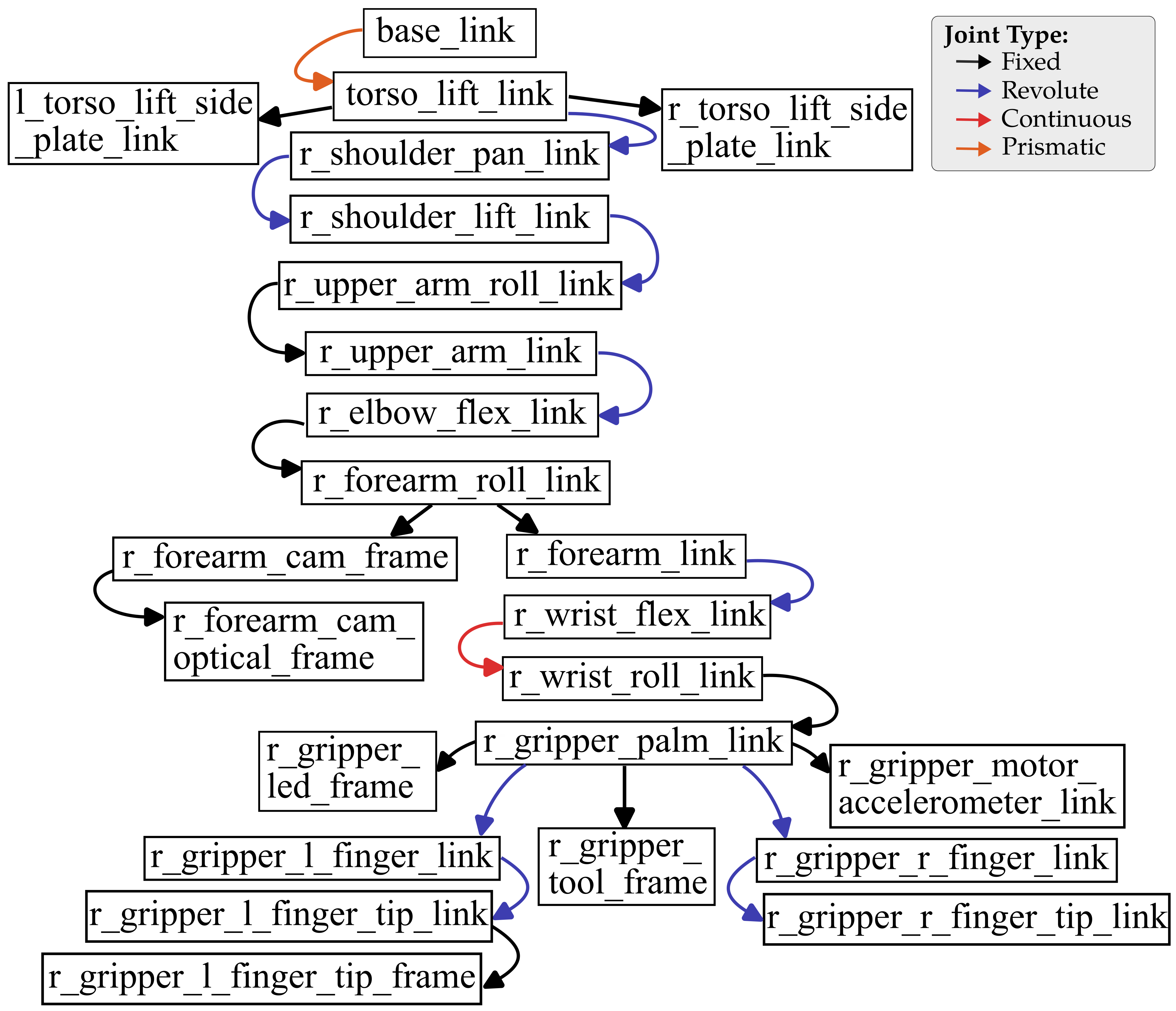}
\caption{Kinematic link tree of the PR2 right arm extracted from its URDF topology. Directed edges represent parent-to-child joint transformations, color-coded by joint type (joint identifiers omitted for visual clarity).}
\label{fig:pr2_arm_urdf}
\end{figure}

To ground our methodology, we employ the right arm sub-tree of the Willow Garage PR2 as a running example throughout the following sections. The stripped kinematic graph spanning from the robot base link to the terminal frames is illustrated in Fig.~\ref{fig:pr2_arm_urdf}. 
While full-scale evaluation is presented in the evaluation section, isolating this articulated arm provides a compact yet representative example that encompasses all key topological modalities:

\begin{itemize}
  \item \textbf{Sensors:} A forearm-mounted camera and a gripper-mounted accelerometer.
  \item \textbf{End-Effector:} A 1-DOF parallel-jaw gripper with a virtual tool frame link.
  \item \textbf{Body Part:} A 7-DOF kinematic chain spanning from \texttt{r\_shoulder} to \texttt{r\_wrist}, attached to the torso.
\end{itemize}

This focused sub-tree provides a clear visual baseline to demonstrate how multi-pass purging progressively simplifies complex kinematic trees towards full-robot abstraction.

\subsection{Neurosymbolic Guardrails}

Rather than relying on unstructured zero-shot prompts, each step of our framework requiring LLM reasoning leverages three standardized, cross-cutting neurosymbolic guardrails:

\paragraph{Majority Voting:}
To balance exploration with output reliability, the framework employs stochastic sampling paired with a majority voting strategy. Because semantic interpretations of ambiguous link/joint naming conventions exhibit variance across inference calls, we enforce output consistency by executing $K$ queries at temperature $T\!>\!0$. The final output is selected via majority consensus. Queries failing to reach a defined consensus threshold $\tau$ (e.g., $\tau \ge 0.6$) can be flagged for human-in-the-loop verification.

\paragraph{Context-Aware Batching:}
Resource-constrained or local LLMs often suffer from attention degradation when prompt length significantly exceeds instruction length~\cite{liu2024lost}. Serializing the entire  $k^{\text{th}}$ pass graph $\mathcal{G}_{\text{URDF}}^{(k)}$ into a single prompt introduces severe context noise. To prevent context degradation, our batching strategy extracts a localized topological subgraph $\mathcal{G}_{\text{sub}} \subset \mathcal{G}_{\text{URDF}}^{(k)}$ around target candidate node $v_i \in \mathcal{V}_L$. This isolates relevant spatial and topological parent-child linkages while bounding token consumption.

\paragraph{Iterative Hierarchical Classification:}
Providing an LLM with either a flattened list of ontology classes or a complex multi-level taxonomy in a single prompt degrades classification accuracy due to context overloading~\cite{wang2022hpt}. To prevent hallucinated concept assignments, taxonomy alignment is executed via top-down constrained classification over the user-provided TBox $\mathcal{T}$. Given a candidate link or sub-tree, the LLM is queried sequentially across depth levels of the taxonomy space $\mathcal{C}$:
\begin{equation*}
\mathcal{C}^{(0)} \supset \mathcal{C}^{(1)} \supset \dots \supset \mathcal{C}^{(d)}
\end{equation*}
At depth $d$, the candidate search space $\mathcal{C}^{(d)}$ is strictly restricted to direct child sub-concepts of the concept validated at depth $d-1$, guaranteeing alignment with the schema.

\section{URDF2Ontology Pipeline}

\begin{figure}[!t]
\centering
\includegraphics[width=\columnwidth]{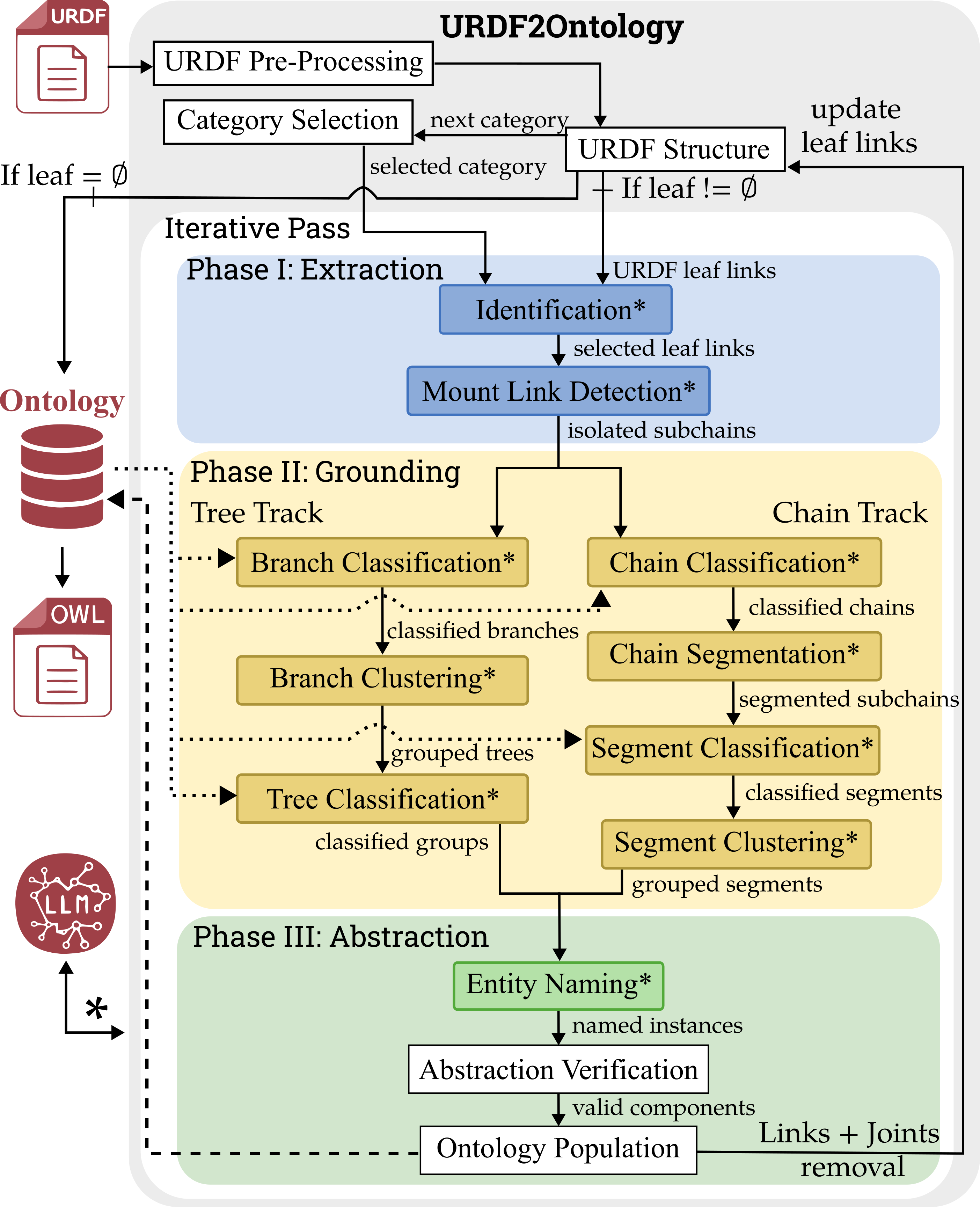}
\caption{\textbf{Overview of the \texttt{URDF2Ontology} pipeline}, tracing execution from raw URDF input to populated ABox assertions. A multi-pass Outer Loop (gray) coordinates the Inner Loop (white), decomposed into three phases: \textbf{Phase I} (blue) isolates candidate sub-chains; \textbf{Phase II} (yellow) grounds semantics via bottom-up assembly (\textit{Tree Track}) and top-down decomposition (\textit{Chain Track}) using the ontology taxonomy; and \textbf{Phase III} (green) handles entity naming, topology verification, and ABox population. Asterisks ($*$) mark LLM-guided steps. Legend: $\rightarrow$ algorithm path, $\dotted$ concept query, $\smalldash$ class/relation assertions.}
\label{fig:pipeline}
\end{figure}

While the Outer Loop governs the global multi-pass workflow, each pass $k$ executes an internal 3-phase pipeline. As illustrated in Fig.~\ref{fig:pipeline}, the Inner Loop transforms the current stripped kinematic graph $\mathcal{G}_{\text{URDF}}^{(k-1)}$ into verified ABox assertions, concluding with active graph purging.

\subsection{Phase I: Extraction}
\label{sec:extraction}

Phase I isolates target component sub-graphs matching the active concept category of pass $k$ with a two-step process:

\paragraph{Step 1: Identification.} 
The framework extracts all current leaf link nodes $\mathcal{V}_{\text{leaf}} \subset \mathcal{V}_L^{(k-1)}$ from the active stripped graph $\mathcal{G}_{\text{URDF}}^{(k-1)}$. The LLM is queried to evaluate these leaf links against the target concept category of pass $k$ (e.g., filtering for sensor frames during Pass 1). Leaves matching the target domain are retained as seed terminal frames $\mathcal{V}_{\text{target\_leaf}}$.

\paragraph{Step 2: Mount Link Detection.} 
For each identified seed leaf $v_{\text{leaf}} \in \mathcal{V}_{\text{target\_leaf}}$, the framework traces its complete kinematic path upward to the root link of the robot tree:

\begin{equation*}
\mathcal{P}(v_{\text{leaf}}) = (l_{\text{leaf}}, j_{\text{leaf}}, l_1, j_1, \dots, l_{\text{root}})
\end{equation*}

This kinematic paths leverage joint information (name and type) since those elements complete the topological structure through actuation information, while not cluttering the LLM's context thanks to the batching mechanism.

Because this complete path includes upstream structural links belonging to the broader robot body, the framework tasks the LLM with identifying the specific boundary node along $\mathcal{P}(v_{\text{leaf}})$ that serves as the \textbf{mount link}. The mount link represents the external structural interface onto which the component is attached. All intermediate links toward $v_{\text{leaf}}$ (excluding the mount link) are classified as constituent parts of the component assembly itself. Trimming all links above the detected mount link yields the isolated candidate sub-graphs $\mathcal{G}_{\text{cand}} \subset \mathcal{G}_{\text{URDF}}^{(k-1)}$ for semantic grounding in Phase II.

\paragraph{PR2 Walkthrough Example.} 
During Pass 1 (Sensors) on the PR2 arm, \texttt{r\_forearm\_cam\_optical\_frame} is identified as a sensor leaf by the LLM.
The framework constructs the ancestor chain and the LLM designates \texttt{r\_forearm\_roll\_link} as the mounting interface. Consequently, the intermediate link \texttt{r\_forearm\_cam\_frame} is recognized as belonging to the sensor assembly, successfully isolating the sub-chain for Phase II without over-trimming or retaining unrelated arm links.

\subsection{Phase II.A: Tree Track Grounding}
\label{sec:tree_track}

The \textbf{Tree Track} is executed during passes dedicated to branching or parallel assemblies, specifically \textbf{Sensors}, \textbf{End-Effectors}, and \textbf{Locomotion}. Because these components frequently consist of heterogeneous multi-branch sub-trees sharing common structural mounts, Phase II.A operates in a \textbf{bottom-up assembly} manner across three sequential steps:

\paragraph{Step 1: Branch Classification.}
For each candidate sub-graph $\mathcal{G}_{\text{cand}}$ isolated in Phase I, using \textit{Iterative Hierarchical Classification}, the LLM is queried with the provided TBox to map each sub-graph to its fine-grained sub-concept.

\paragraph{Step 2: Branch Clustering.}
Because multiple independently extracted sub-graphs may belong to the same physical unit (e.g., left and right fingers of a single gripper, or multiple optical frames of the same sensor), \textit{Branch Clustering} provides the LLM with all validated branch sub-graphs from Phase I simultaneously. The LLM is tasked to evaluate their kinematic topology to group co-located branches into unified \textbf{assembly units}.

\paragraph{Step 3: Tree Classification.}
Finally, for each clustered assembly unit, the LLM executes a top-down \textit{Iterative Hierarchical Classification} on the combined sub-tree structures. By leveraging the aggregated semantic context of all constituent branches, the LLM determines the composite entity class $C \in \mathcal{C}$ defined in TBox $\mathcal{T}$.

\paragraph{PR2 Walkthrough Example.}
To illustrate Phase II.A on the PR2 right end-effector:
\begin{enumerate}
  \item \textbf{Branch Classification:} Phase I isolates three sub-graphs mounted on \texttt{r\_gripper\_palm\_link}: \texttt{r\_gripper\_r\_finger\_tip\_link}, \texttt{r\_gripper\_l\_finger\_tip\_frame}, and \texttt{r\_gripper\_tool\_frame}. The LLM classifies the first two as \texttt{Finger} and the third as a \texttt{ToolCenterPoint}.
  \item \textbf{Branch Clustering:} The LLM evaluates all three sub-graphs together. Recognizing their functional similarity and shared connection to \texttt{r\_gripper\_palm\_link}, it clusters them into a single end-effector assembly unit.
  \item \textbf{Tree Classification:} Using the input TBox $\mathcal{T}$ with the full assembly unit (comprising two \texttt{Finger} and one \texttt{ToolCenterPoint}), the LLM classifies the overall clustered sub-tree as a \texttt{TwoFingerGripper}.
\end{enumerate}

\textit{Key Insight:} Bottom-up aggregation is essential here because high-level composite concepts cannot be reliably inferred from a single link alone. Observing two opposing \texttt{Finger} links alongside a \texttt{ToolCenterPoint} provides the structural context required to classify the assembly as a \texttt{TwoFingerGripper}. In addition, retaining branch-level classifications throughout the pipeline yields a fine-grained description in the final knowledge base. Rather than collapsing the sub-tree into a single monolithic label, both the macro-level assembly unit and its sub-components will be instantiated and linked in the target ABox.


\subsection{Phase II.B: Chain Track Grounding}
\label{sec:chain_track}

While the Tree Track addresses branching sub-trees via bottom-up assembly, during Pass 4+ the \textbf{Chain Track} processes body parts (e.g., arms, head, torsos), consisting of serial kinematic paths. 
Phase II.B takes the isolated candidate sub-graphs $\mathcal{G}_{\text{cand}}$ extracted from Phase I (now representing continuous kinematic chains freed from peripheral sensors and end-effectors by prior purge passes) and grounds them via a \textbf{top-down decomposition} strategy across four steps:

\paragraph{Step 1: Chain Classification.}
Each isolated chain $\mathcal{G}_{\text{cand}}$ is first evaluated as a single macro-level entity. The LLM performs \textit{Iterative Hierarchical Classification} against the user-provided TBox $\mathcal{T}$ to assign a top-level category to the overall chain (e.g., classifying a chain as an \texttt{Arm}).

\paragraph{Step 2: Chain Segmentation.}
Having established the macro-level context, the framework tasks the LLM with splitting the continuous chain into candidate contiguous sub-segments. The LLM can analyze the joint types and link naming patterns along the chain to identify natural functional transition points without requiring external heuristics.

\paragraph{Step 3: Segment Classification.}
Each newly isolated segment is independently classified via \textit{Iterative Hierarchical Classification}. The LLM maps each segment to its specific sub-anatomical concept (e.g., \texttt{UpperArm}, \texttt{Forearm}).

\paragraph{Step 4: Segment Clustering.}
To ensure robustness against LLM over-segmentation or non-standard URDF modeling conventions, a final verification stage is executed. The LLM evaluates all classified segments to group adjacent or parallel sub-segments of the same type that collectively represent a physical body part. This step aims at consolidating multi-link sub-assemblies or resolving parallel topologies (e.g., dual prismatic joints acting as a single linear actuator).

\paragraph{PR2 Walkthrough Example.}
To demonstrate Phase II.B on the PR2 right arm:
\begin{enumerate}
  \item \textbf{Input Chain:} Prior passes (Sensors and End-Effectors) have already abstracted and purged peripheral devices. Phase II.B receives a continuous chain from \texttt{r\_wrist\_roll\_link} to \texttt{r\_shoulder\_pan\_link}.
  \item \textbf{Chain Classification:} The LLM evaluates the entire kinematic chain and classifies it as an \texttt{Arm}.
  \item \textbf{Chain Segmentation:} The LLM decomposes the \texttt{Arm} chain into 5 segments based on joints and naming.
  \item \textbf{Segment Classification:} Each segment is classified against TBox $\mathcal{T}$, yielding: \texttt{Shoulder}, \texttt{UpperArm}, \texttt{Elbow}, \texttt{ForeArm}, and \texttt{Wrist}.
  \item \textbf{Segment Clustering:} The LLM inspects the 5 classified segments for over-segmentation or parallel redundancies. In this baseline PR2 case, the 5 segments correspond 1:1 with distinct functional units, so no further merging is required, validating the final segmented structure.
\end{enumerate}

\textit{Key Insight:} Top-down decomposition resolves a fundamental challenge in robot parsing: intermediate links in serial chains often lack explicit semantic identifiers. By first establishing the macro-level context, the LLM can leverage global topological position to reliably segment and classify ambiguous intermediate components. Furthermore, combining macro-chain and micro-segment individuals in the ABox provides a complete, multi-tiered hierarchy in the final knowledge base.


\subsection{Phase III: Abstraction \& Graph Purging}

Following identification, grounding, and clustering/segmentation, Phase III executes the final verification, symbolic instantiation, and graph modification steps prior to initiating the next outer pass.

\paragraph{Step 1: Entity Naming.}
Because Phases II.A and II.B aggregate multi-branch sub-trees and segment continuous kinematic paths into composite units, the naive one-to-one mapping between raw URDF link identifiers ($v \in \mathcal{V}_L$) and ontology individuals no longer holds. The \textit{Entity Naming} step use the LLM to generate unique, semantically explicit URI/individual identifiers $i \in \mathcal{I}$ for both top-level composite assemblies (e.g., \texttt{pr2\_right\_arm}) and their constituent sub-components or segments (e.g., \texttt{pr2\_right\_forearm}).

\paragraph{Step 2: Abstraction Verification.}
Before committing instances to the knowledge base, the framework evaluates the topological integrity of the candidate sub-graph relative to the global robot tree. Specifically, it verifies that extracting the target sub-graph will not create disconnected ("orphan") kinematic links elsewhere in $\mathcal{G}_{\text{URDF}}^{(k-1)}$. If a structural anomaly or potential orphan sub-tree is detected during verification, the extracted entity is still pushed to the ontology but annotated with a dedicated \textit{OrphanLink} quality flag in the ABox. This ensures system execution remains fully automated while flagging uncertain topological states for future human-in-the-loop audit.

\paragraph{Step 3: Ontology Population.}
Once verified, the \textit{Ontology Population} step commits the grounded entities into the target ABox according to the schema provided in TBox $\mathcal{T}$. First, named individuals $i \in \mathcal{I}$ are created for macro-assemblies and sub-parts. Then, concept types are asserted for each individual ($C(i)$, where $C \in \mathcal{C}$). Finally, object properties ($R(i_1, i_2)$, where $R \in \mathcal{R}$) establish structural and hierarchical ties, such as $\textit{hasSubPart}(i_{\text{arm}}, i_{\text{forearm}})$ or $\textit{isMountedOn}(i_{\text{camera}}, i_{\text{arm\_link}})$.

\begin{figure}[!h]
\centering
\includegraphics[width=\columnwidth]{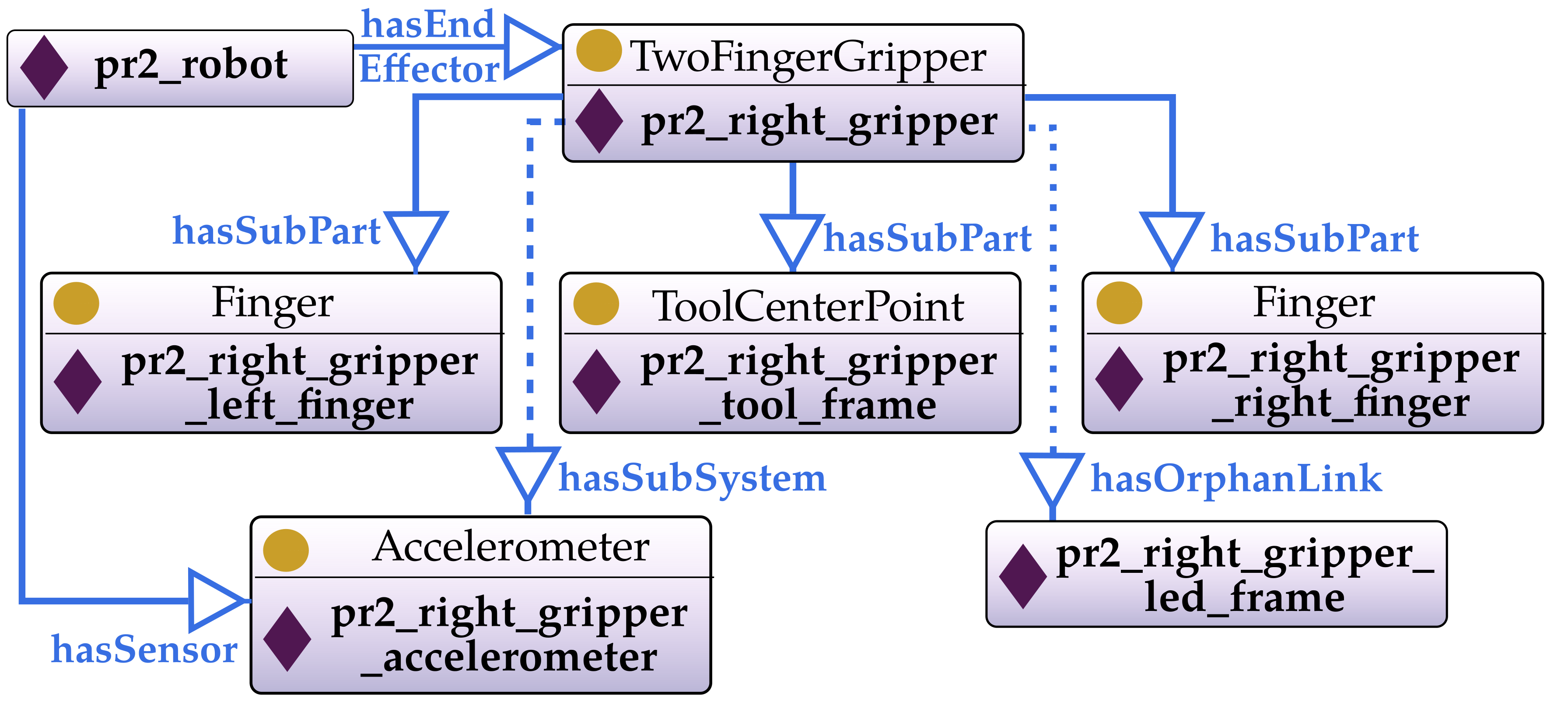}
\caption{Populated ontology graph for the PR2 gripper semantic abstraction, where \raisebox{-0.2\height}{\includegraphics[height=1em]{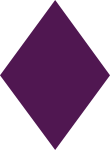}} indicates named individuals, \raisebox{-0.2\height}{\includegraphics[height=1em]{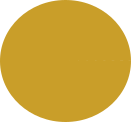}} a concept class, and \raisebox{-0.2\height}{\includegraphics[height=1em]{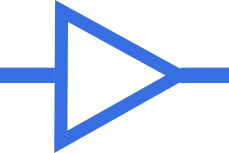}} an object property relation. Dotted edges (\raisebox{-0.2\height}{\includegraphics[height=1em]{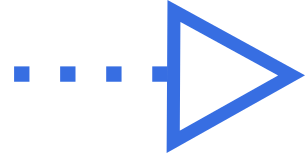}}) connect distinct components (e.g., gripper and accelerometer), while dashed edges (\raisebox{-0.2\height}{\includegraphics[height=1em]{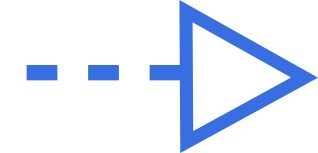}}) denote links to orphan nodes. }\label{fig:ontology_gripper}
\end{figure}

\paragraph{PR2 Walkthrough Example.}
To illustrate the output of Phase III, consider the populated ABox sub-graph generated for the PR2 right end-effector at the end of Pass 2 (Fig.~\ref{fig:ontology_gripper}):
\begin{enumerate}
  \item \textbf{Entity Naming \& Sub-part Assertions:} The framework generates a macro \texttt{TwoFingerGripper} individual (\texttt{pr2\_right\_gripper}). Through \textit{Ontology Population}, three constituent sub-parts are instantiated and linked to it via hasSubPart object properties: two \texttt{Finger} individuals and one \texttt{ToolCenterPoint} for the virtual tool frame.
  \item \textbf{Cross-Pass Subsystem Integration:} The \texttt{Accelerometer} individual previously identified, grounded, and asserted during Pass 1 (Sensors) is linked to the gripper assembly via an hasSubsystem object property. This demonstrates how sequential purging and ABox assertion maintain long-term functional relationships across different outer passes.
  \item \textbf{Global Platform Aggregation:} Upon each pass, newly instantiated macro components are connected back to a main robot individual (\texttt{pr2\_robot}) via category-specific object properties (e.g., \texttt{hasSensor}). Redirecting all functional entities back to this central individual facilitates manual verification and streamlines interfacing with Description Logic reasoning patterns for capability inference.
\end{enumerate}

\paragraph{Post-Phase Purging Operator.}
Upon completing ABox population for pass $k$, all link and joint entities belonging to the abstracted sub-graph $\mathcal{G}_{\text{cand}}$ are purged from the URDF structure.
By doing so, the active graph is reduced ($|\mathcal{V}_L^{(k)}| < |\mathcal{V}_L^{(k-1)}|$), yielding a simplified, context-isolated kinematic tree $\mathcal{G}_{\text{URDF}}^{(k)}$ ready for the subsequent outer pass.


\section{Experimental Evaluation}

To evaluate framework performance across heterogeneous fleet topologies, we compiled a benchmark dataset of aerial, wheeled, and humanoid URDF models sourced from public model repository. Incorporating lightweight aerial platforms alongside complex, multi-limb humanoids ensures coverage across varying kinematic complexities. Evaluation was performed against a custom-designed OWL ontology that defines a formal taxonomy for high-level hardware categories and their subcomponents, spanning sensors, end-effectors, locomotion systems, and body parts.

We evaluate our framework using a local, open-weight LLM (\texttt{gpt-oss:20b}) via Ollama, selected for its strong reasoning performance and rapid prompt processing speed. Because target output schemas cannot be natively enforced during generation with reasoning-capable models, we specify the expected format within the task prompt and perform post-hoc syntactic verification on every model response. All benchmarks were conducted on an Ubuntu 24.04 LTS workstation equipped with an Intel Core i7-12850HX CPU (24 cores), 32 GB RAM, and an NVIDIA RTX A4500 Laptop GPU (16 GB VRAM).

Our \texttt{URDF2Ontology} code, selected URDFs, target taxonomy, and resulting populated ontologies are available at the top of the paper.

\subsection{Ontology Mapping Accuracy}

To evaluate the generalization and structural extraction capabilities of \texttt{URDF2Ontology}, we selected five representative robotic platforms spanning distinct kinematic topologies, sensor density levels, and naming conventions.
In the following we discuss for each robot the quality of the generated ontologies in terms of both strengths and weaknesses. A quantitative distribution of extracted structural entities and grounded functional concepts is provided in Table~\ref{tab:concept_assertion_results}.

\paragraph{Solo (Quadruped):} The Solo quadruped evaluates semantic extraction under implicit, highly abbreviated naming conventions within multi-legged kinematic, compressing 13 raw URDF links into 5 ontology individuals. Despite the absence of onboard sensors or articulated end-effectors, \texttt{URDF2Ontology} correctly identified four distinct \texttt{ArticulatedLeg} components and anchored the assembly under a \texttt{QuadrupedalBase}. Crucially, the framework successfully resolved cryptic, domain-specific link identifiers (such as \texttt{HL\_KFE} and \texttt{HL\_HFE}) into human-interpretable instance names (e.g., \texttt{hind\_left\_lower\_foot} and \texttt{hind\_right\_upper\_foot}). This case study also illustrates a key architectural trade-off of our design: because multi-branching locomotion topologies are processed via the bottom-up \textit{Tree Track}, intermediate limb links are aggregated into holistic functional individuals rather than expanded into granular serial link sub-components (as would occur in the top-down \textit{Chain Track}). Consequently, instance labeling favors terminal link semantics (e.g., foot) instead of high level functionality.

\paragraph{RB-VOGUI+ (Composite Mobile Manipulator):} The RB-VOGUI+ platform exemplifies composite robotic systems by integrating an all-terrain wheeled base with a top-mounted robotic arm, abstracting 53 raw URDF links into 34 ontology individuals. Despite the platform's architectural heterogeneity, \texttt{URDF2Ontology} successfully decoupled the primary subsystems into a four-wheeled mobile base (\texttt{DifferentialDriveBase}) and a dedicated \texttt{SerialManipulatorArm}. Additionally, the arm exposes a common edge case in URDF modeling conventions: rather than a single continuous linear chain, the arm's kinematic tree is split across two parallel branches. Treating these sub-chains independently would incorrectly instantiate two disconnected \texttt{Arm} individuals. This scenario highlights the necessity of the \textit{Segment Clustering} phase within the \textit{Chain Track}, which successfully merged both branches into a single \texttt{Arm} entity.

\paragraph{PR2 (Mobile Manipulator):} The PR2 mobile manipulator serves as a benchmark for complex, multi-assembly platforms, abstracting 82 raw URDF links into 61 formal ontology individuals. The framework successfully identified and grounded the platform's primary sub-systems: the articulated head was classified as a \texttt{PanTiltHead}, while both 7-DOF arms were parsed with high segment-level granularity into distinct \texttt{SerialManipulatorArm} instances terminating in fine-grained \texttt{EndEffector} assemblies. The platform's locomotion system highlights the framework's single-level assembly abstraction boundaries. PR2 utilizes four articulated casters, each hosting a dual-wheel setup (eight wheels in total). The framework grounded these structures at the intermediate \texttt{WheeledLocomotion} level rather than decomposing individual wheels. This behavior represents a logical trade-off as bypassing the caster level to group all wheels directly would yield an overly broad base abstraction, whereas preserving the caster node accurately captures the functional locomotion unit.
Finally, even if the tested taxonomy proposed a \texttt{Lidar} concept, the framework classified the two onboard LIDARs as \texttt{LaserRangeFinder} as their raw link identifiers containing only the generic token \texttt{laser}.

\paragraph{iCub (Dense Humanoid):} The iCub humanoid platform serves as an empirical stress test for ultra-dense multi-limb architectures, featuring dual 5-finger hands, bipedal legs, and extensive sensor networks. Significantly, an initial trial incorporating the robot's ~150 raw tactile skin sensor cells led to prompt degradation during the unbatched grouping phase, triggering severe LLM hallucinations (spelling out raw code output). This failure empirically validates our core thesis: monolithic LLMs cannot digest un-decomposed, ultra-dense URDF files without context isolation. Filtering these micro-skin cells, \texttt{URDF2Ontology} evaluated the remaining platform, abstracting 134 raw URDF links into 84 ontology individuals. The framework successfully isolated 30 distinct sensor-related frames, grouping them into 19 functional sensor entities. Additionally, the 10 micro-kinematic finger sub-chains (5 per limb) were accurately recognized and aggregated into two cohesive \texttt{DexterousHand} assemblies. Finally, both bipedal legs were correctly identified. However, consistent with the \textit{Solo} evaluation, internal leg segment granularity was aggregated into macro-limb concepts due to exclusive \textit{TreeTrack} processing in Pass~3.

\paragraph{Ingenuity (Aerial Vehicle):} The Ingenuity helicopter provides an edge-case evaluation for aerial topologies, successfully abstracting 13 raw URDF links into 9 ontology individuals. The dual rotor blades were correctly grouped and asserted into a unified \texttt{RotorAssembly}. Additionally, because each of the four landing legs contains a revolute joint, the framework classified them as \texttt{ArticulatedLeg} components and assigned the overall assembly to a \texttt{QuadrupedalBase}. Despite no cues regarding joint actuation (i.e., distinguishing active motors from passive springs), the framework accurately captures physical morphology purely through kinematic topology. This result highlights the framework's versatility in supporting non-standard and multi-modal locomotion structures.

\begin{table*}[htbp]
\centering
  \begin{tabular}{|l|c|c||c|c|c|c|}
  \hline
  \textbf{\begin{tabular}[c]{@{}c@{}}Robot\\ Model\end{tabular}} &
    \textbf{\begin{tabular}[c]{@{}c@{}}Sub-part \\ Segment\end{tabular}} &
    \textbf{\begin{tabular}[c]{@{}c@{}}Assembly \\ Chain\end{tabular}} &
    \textbf{Sensor} &
    \textbf{End-effector} &
    \textbf{\begin{tabular}[c]{@{}c@{}}Locomotion\\ System\end{tabular}} &
    \textbf{Body-part} \\ \hline\hline
  Solo      & 4  & 1  & 0  & 0 & 1 & 0 \\ \hline
  RB-Vogui+ & 25 & 9  & 6  & 1 & 1 & 2 \\ \hline
  Pr2       & 43 & 18 & 9  & 2 & 4 & 4 \\ \hline
  iCub      & 47 & 31 & 19 & 2 & 2 & 6 \\ \hline
  Ingenuity & 7  & 2  & 1  & 0 & 2 & 0 \\ \hline
  \end{tabular}
  
  \caption{Quantitative distribution of extracted structural entities and grounded functional concepts. Note that functional assertions encompass both multi-link assemblies and standalone links, causing concept totals to reflect both structural granularities.}
\label{tab:concept_assertion_results}
\end{table*}

\section{Discussion \& Limitations}

While our empirical evaluation is bounded to five representative robot platforms, a single target domain TBox, and local open-weight LLMs, performance metrics may naturally vary across alternative model architectures, user taxonomies, or URDF modeling conventions. Crucially, however, attempting to process raw, undecomposed URDF trees directly presents a fundamental bottleneck: as link density and kinematic depth increase, the raw string representation rapidly saturates the LLM context window, triggering reasoning degradation and topological hallucinations. By isolating context through sequential graph purging and enforcing schema constraints, \texttt{URDF2Ontology} renders semantic embodiment grounding tractable, even for constrained local models. Nevertheless, our evaluation highlights several specific limitations of the current framework:

\paragraph{Dependence on Identifier Semantics:} While the framework leverages structural topology (parent-child links, joint types) to guide classification, it primarily relies on semantic cues embedded in link identifiers. Much like human annotators, the framework fails to correctly interpret highly obfuscated URDFs or purely numerical naming conventions (e.g., \texttt{link\_01}, \texttt{node\_a}).

\paragraph{Schema Bounds and Taxonomy Granularity:}
Because \texttt{URDF2Ontology} operates against a static target TBox, platform devices lacking an explicit representation in the input schema risk misclassification or, more frequently, falling back to overly broad high-level ancestor concepts. While dynamic taxonomy enrichment offers a promising future direction, the current system remains bounded by the predefined schema. Furthermore, to prevent evaluation bias from hand-crafted metadata, our classification tasks rely exclusively on class URIs, omitting rich annotation properties such as \texttt{rdfs:comment} or \texttt{skos:definition} that could clarify ambiguous concepts. Finally, because iterative classification evaluates taxonomy depth strictly one layer at a time to minimize token consumption, the model lacks visibility into deeper sub-concepts when making high-level decisions. While these constraints guarantee minimal token overhead and strict benchmark fairness, incorporating annotation-aware prompts and broader taxonomy context represents a key trade-off for schema alignment.

\paragraph{Locomotion Topology Granularity:}
Currently, Pass~3 processes locomotion systems strictly through the \textit{Tree Track}. While this strategy reliably captures macro-level mobility platforms (e.g., \texttt{QuadrupedalBase}, \texttt{DifferentialDriveBase}), it introduces a topological mismatch on legged systems. Legged platforms exhibit a hybrid kinematic structure, combining high-level branching limbs with low-level serial joint chains within each leg. By processing locomotion exclusively as a tree topology, intermediate limb sub-components (e.g., \texttt{Thigh}, \texttt{Shank}) are aggregated into top-level macro-assemblies prior to graph purging, suppressing fine-grained anatomical entities. To resolve this, future extensions to Pass~3 need to incorporate a \textit{Hybrid Grounding Strategy} that unifies macro-level assembly clustering with micro-level chain decomposition, leveraging top-level limb context to segment serial sub-structures before graph purging.

\paragraph{Dynamic Reasoner Feedback:}
While the framework actively leverages the target TBox to guide its extraction and classification, the current execution pipeline operates without dynamic feedback from formal Description Logic reasoners (e.g., HermiT, Pellet). As a result, if an LLM output causes the generation of assertions that violate underlying TBox logical axioms, such as class disjointness or domain/range restrictions, the framework cannot dynamically detect or repair these inconsistencies at runtime. Closing this loop by incorporating real-time reasoner consistency checks to trigger self-correcting prompts represents a natural next step for the architecture.

\section{Conclusion}

In this work, we presented \texttt{URDF2Ontology}, a neurosymbolic framework that automatically transforms raw URDF files into populated ontologies. By introducing an iterative, sequential graph-purging mechanism, our architecture eliminates LLM context window saturation while scaling effectively to complex, multi-branching robot topologies. Through dual-track grounding, combining bottom-up assembly for peripheral devices with top-down decomposition for continuous chains, our framework generates aligned assertions against arbitrary user-provided TBox schemas.

Experimental evaluations across five diverse robotic embodiments (a mobile manipulator, quadruped, humanoid, wheeled vehicle, and aerial vehicle) demonstrate that our approach significantly reduces manual knowledge-engineering overhead while guaranteeing strict vocabulary alignment with the target ontology.

In future work, we plan to integrate automated OWL reasoner validation and human-in-the-loop feedback directly into the inner loop to enable real-time, self-correcting ontology population. Additionally, we aim to incorporate extra pipeline modules, such as Gazebo plugin parsers, to extract richer semantic properties (e.g., sensor parameters, gripper opening widths). Finally, automatically generating standard SRDF files directly from the abstracted ontology will help bridge the gap between low-level kinematic specifications and high-level motion-planning frameworks.

\section{Acknowledgments}
This work was supported by the French National Research Agency (ANR) under project HumFleet (grant ANR-23-CE33-0003) and project ELSA (grant ANR-21-CE33-0019)

\bibliography{dussard_camera_ready}

\end{document}